\documentclass[journal]{IEEEtran}
\usepackage{cite}
\usepackage{amsmath,amssymb,amsfonts}
\usepackage{algorithmic}
\usepackage{graphicx}
\usepackage{textcomp}
\usepackage{xcolor}
\usepackage{algorithm}
\usepackage[normalem]{ulem}
\usepackage{booktabs}       
\usepackage{hyperref}

\ifCLASSINFOpdf
\else
\fi
\begin{document}
%
\title{Contextually Enhanced ES-dRNN with Dynamic Attention for Short-Term Load Forecasting}
%
%
%

\author{Slawek~Smyl,
	Grzegorz~Dudek,
	and~Paweł~Pełka

\thanks{S. Smyl works at Meta, 1 Hacker Way, Menlo Park, CA 94025, USA,
e-mail: slawek.smyl@gmail.com}
\thanks{G. Dudek and P. Pełka are with the Department of Electrical Engineering, Czestochowa University of Technology, 42-200 Czestochowa, Al. Armii Krajowej 17, Poland, e-mail: grzegorz.dudek@pcz.pl, pawel.pelka@pcz.pl.}
}

\maketitle

\begin{abstract}
In this paper, we propose a new short-term load forecasting (STLF) model based on contextually enhanced hybrid and hierarchical architecture combining exponential smoothing (ES) and a recurrent neural network (RNN). The model is composed of two simultaneously trained tracks: the context track and the main track. The context track introduces additional information to the main track. It is extracted from representative series and dynamically modulated to adjust to the individual series forecasted by the main track. The RNN architecture consists of multiple recurrent layers stacked with hierarchical dilations and equipped with recently proposed attentive dilated recurrent cells. These cells enable the model to capture short-term, long-term and seasonal dependencies across time series as well as to weight dynamically the input information. The model produces both point forecasts and predictive intervals. The experimental part of the work performed on 35 forecasting problems shows that the proposed model outperforms in terms of accuracy its predecessor as well as standard statistical models and state-of-the-art machine learning models.  

\end{abstract}
 

%
\IEEEpeerreviewmaketitle

\section{Introduction}
\IEEEPARstart{E}{}lectricity demand forecasting is extremely important for the energy sector to ensure the secure, effective and economic operation of the power system. 
The need for forecasting results from the fact that electricity cannot be stored on a massive scale, so the supply must be adequate to meet all customers’ demand simultaneously, instantaneously and reliably. To provide electricity on demand, electric system operations have to be planned and conducted with those goals in mind.

Long- and medium-term load forecasting predict the power system load over time ranges measured in months or years, while short-term load forecasting (STLF) concerns a forecast horizon of a few hours to a few days. STLF is necessary to schedule generation resources to meet the future load demand and to optimize the power flow on the transmission network to reduce congestion and avoid overloads. It is used for unit commitment, generation dispatch, hydro scheduling, hydrothermal coordination, spinning reserve allocation, interchange and low flow evaluation, fuel allocation and network diagnosis. Improving the accuracy of forecasting can significantly reduce the operating costs of the power system and increase its security. As electricity pricing is driven by supply and demand, STLF plays a key role in competitive electricity markets. It determines the financial performance of the market participants such as generation and transmission utilities, electricity traders and financial institutions. Accurate forecasts translate directly into reduced financial risks and improved financial performance.

STLF is a difficult problem due to the complexity of electricity demand time series, which can express nonlinear trend, multiple seasonality (yearly, weekly and daily), variable variance, random fluctuations and changing daily profiles. They are influenced by many stochastic factors such as climatic, weather and economic conditions. The increasing contribution of volatile, fluctuating renewable energy sources, which has been observed in recent years, introduces an additional random component to electricity demand time series. The complexity of the problem and its importance for the safe, reliable and effective operation of power systems and electricity markets have generated great interest among researchers in this field. 

\subsection{Related Work}

STLF complexity places high demands on forecasting models. They have to deal with non-stationarity and multiple seasonality in time series. Roughly, STLF methods can be classified into statistical/econometric and more sophisticated machine learning (ML) methods. The first category has numerous disadvantages, such as linear nature, limited adaptability, limited ability to deal with complex seasonal patterns, problems with introducing exogenous variables into the model and capturing long-term relationships in time series \cite{Smy21}. Typical representatives of this category are: auto-regressive integrated moving average (ARIMA) \cite{Aro18}, exponential smoothing (ES) \cite{Tay12}, linear regression \cite{Cha14}, and Kalman filtering \cite{Tak16}. 

ML and computational intelligence methods are more flexible in modeling complicated nonlinear relationships than statistical approaches. They include well-known ML methods such as classical neural networks (NNs) \cite{Dud16a}, tree-based models \cite{Apr21} , support vector machine \cite{Li20}, neuro-fuzzy systems \cite{Kan22} as well as the latest developments in the field of deep learning \cite{Hon20}, recurrent NNs \cite{Kon19}, boosted decision trees \cite{Wan21}, hybrid and ensemble models \cite{Mas21, Tan20}. 

In recent years, the forecasting literature has mainly focused on NNs, especially deep, recurrent and hybrid solutions, but also tree-based models should be mentioned due to their high performance in forecasting competitions \cite{Boj21}.
The success of deep NNs (DNNs) can be largely attributed to their increased complexity which allows the model to learn the appropriate data representation for modeling the target function. While the success of RNNs is due to their ability to capture complex, short- and long-term temporal relationships in time series and thus model the dynamics of the series more accurately. 
New ideas and achievements in deep learning can be immediately assimilated by STLF models. Some examples of STLF models employing state-of-the-art RNN and DNN are: \cite{Esk21}, where convolutional NNs (CNNs) were used to extract load and temperature features that feed the bidirectional propagating RNN to make hourly electrical load forecasts; \cite{Shi18}, where pooling-based deep learning is proposed, which batches a group of customers' load profiles into a pool of inputs and is able to learn the uncertainty of load profiles;  \cite{Kim19}, where a recurrent inception CNN is proposed for STLF that combines RNN and 1-dimensional CNN;  \cite{Wan19}, where a bi-directional long short-term memory NNs with attention mechanism is proposed; \cite{Che19}, where deep residual networks integrate domain knowledge and researchers' understanding of the problem and produce probabilistic load forecasts using Monte Carlo dropout; and \cite{Lin21}, where an attentive transfer learning based graph NN is proposed that can utilize the learned prior 
knowledge to improve the learning process for residential electric load forecasting.

Many of the above mentioned STLF models use ensembling or hybridization to increase forecast accuracy and robustness. Combining base models in the ensemble can take many forms. For example in \cite{Dud21}, to reduces the model variance related to the stochastic optimization algorithm as well as data and parameter uncertainty, three levels of ensembling were used at the same time: stage of training level, data subset level, and model level. In \cite{Yan22}, an ensemble of RNNs is applied, which learns on the components of the bivariate empirical mode decomposition, to forecast the interval-valued load. 
Another example of an ensemble based on the products of time series decomposition can be found in \cite{Gao22}, where an ensemble of randomized DNN combined with an empirical wavelet transformation is proposed.  In \cite{Pio22}, several ensemble methods aggregating different ML models are compared for one-day-ahead wind power forecasting. 

Hybrid models that combine two or more base models into one take advantages of the best features of each base model while eliminating their drawbacks. The beneficial effects of the model hybridization for STLF were demonstrated in many papers. For example, in the approach proposed in \cite{Smy21}, a statistical method (ES) helps in data preprocessing for ML model (RNN). Namely, ES extracts dynamically the main components of each individual time series and enables appropriate series representation for RNN. This makes RNN capable of dealing with complex series expressing multiple seasonality. 
In \cite{Wan21}, a temporal CNN is utilized to extract hidden information and long-term temporal relationships in the data and a boosted tree model (LightGBM) is applied to predict future loads based on the extracted features.
The approach presented in \cite{Mas21} combines three models: LightBoost, XGBoost and NN. The first two models generate the metadata while NN uses them to calculate the final predictions. 

\subsection{Motivation, Inspiration and Contribution} 

Global ML forecasting models learn from all series in a dataset, however when forecasting a particular series, only inputs derived from this one series are typically used. For every series there is likely to exist one or more series in the dataset, that, if used  as an additional input, would improve the forecasting accuracy. These issues have been researched in depth. In the context of statistical models we have the notion of Granger causality \cite{Gra69}, there are factor models \cite{Satorras21}, \cite{Sen19}, \cite{Wan19a}, \cite{Stefani21} and in ML there has been quite a bit of work done on Graphical NNs \cite{Zhou18}, \cite{Zonghan20}, \cite{Li18}, \cite{Tran21}. In the ML context there are two main issues here: how to find a small subset of "friendly" time series out of many, perhaps dozen or hundreds of thousands of others, and having found them, how to aggregate the features extracted, considering that just concatenating them is unlikely to succeed, because in ML systems inputs are position sensitive. Additionally, if, as is typical, the  number of "friendly" series, or neighborhood size, is a constant, it is a rigid constraint, too low for some series and too high for other series.

We propose a RNN-based forecasting system composed of two tracks: the first one, the context track, processes a representative subset of series and outputs an opaque vector that, optionally after being modulated to suit a particular series, is concatenated with other features and becomes an input to the second, main forecasting track. The idea of context has been inspired by \cite{Qin17}, but the context here is calculated in a quite different, and much more computationally efficient way, notably it does not use standard attention (distinct from a completely different attention mechanism used internally by adRNNCell, see below).

The architecture of each track is inspired by the winning submission to the M4 forecasting competition 2018 \cite{Smy20}. This hybrid and hierarchical forecasting model was composed of ES and RNN modules that are simultaneously trained by the same optimization algorithm.
ES allows for on-the-fly preprocessing of the series, i.e. deseasonalization and adaptive normalization, and thus captures the main features of the individual series. RNN is composed of several stacked long short term memory cells (LSTMs) with dilated and residual connections, which improve long-term memory performance and speed-up training. Due to the hybrid and hierarchical architecture and cross-learning on many series, the model is able to learn both local and global time series features. This and also ensembling at three levels improve forecasting accuracy. Based on this solution, in \cite{Dud21}, we developed a model for mid-term load forecasting and in \cite{Smy21} we developed a model for STLF. The latter model uses a new type of recurrent cell with additional delayed states and an additional output vector. The delayed states help to model seasonalities and long-term relationships, while introducing an additional output vector allows the cell to better control the flow of information. An interesting feature of the model is that RNN learns not only the forecasting function but also the ES parameters to ensure optimal time series preprocessing. A further refinement of the STLF model proposed in \cite{Smy22} is the introduction of an internal attention mechanism that dynamically weights the inputs for the cell.

In continuing the development of the STLF models based on a hybrid and hierarchical architecture combining ES and RNN, this study makes the following contribution: we propose a new STLF model which is composed of two tracks: the context track and the main track. The context track produces a dynamic context vector based on the history, up to the moment, of a representative group of series. This vector is modulated to adjust to the forecast series and introduced as an additional input to the main track RNN. It extends per-series input data with information contained in the context series and thus supports forecasting individual series. The model inherits the features of its predecessors, i.e. attentive dilated recurrent cells, multiple dilated recurrent layers stacked with hierarchical dilations, cross-learning ability, ability to learn the ES smoothing coefficients, and ability to produce both point forecasts and predictive intervals (PIs). 
We provide the open-source code of our model in the github repository \cite{rep22}. In the experimental part of the work covering STLF for 35 European countries, we show that our proposed model is more accurate than its predecessor and 16 baseline models, including both statistical and ML models.

The rest of the paper is structured as follows. First, the forecasting model is described in Section II, which includes the STLF problem, model architecture and components, and loss function. Then, an extensive experimental evaluation is provided in Section III. Finally, conclusions are drawn in Section IV.

\section{Forecasting Model}

In this section we define the forecasting problem and present the proposed solution. We describe the functions of the context and main tracks, the model architecture and its components, and the loss function which enables the model to produce both point forecasts and predictive intervals (PIs).  

\subsection{STLF Problem Statement} 

The goal of STLF is to use the previously observed loads of the power system to forecast a fixed length of the future loads. In this study, we consider hourly load time series and a 24-hour forecast horizon. Thus the goal is to predict $\{{z}_\tau\}_{\tau=M+1}^{M+24}$, given a sequence of past observations, $\{z_\tau\}_{\tau=1}^{M}$, where $z_\tau$ is the system load at hour $\tau$ and $M$ is the time series length (training part). 

This STLF problem is challenging because the load time series expresses a nonlinear trend, varying variance, random fluctuations and triple seasonality: daily, weekly and yearly. The daily load profiles differ in shape depending on the day of the week and season of the year. The load time series nature is specific for different countries. It is shaped by socioeconomic factors such as population, urbanization, industrialization, level of technological development, electricity price as well as climatic and weather conditions. 
To get a broader picture of the load time series characteristics for European countries see \cite{Smy21}, where we analysed load series in terms of variability for various horizons, seasonal components and similarity of daily profiles. 

\subsection{Context Track and Main Track} 

The proposed forecasting model is composed of two tracks: the context track and main forecasting track, see Fig. \ref{figBD}. A helpful context to a model forecasting a series should be one that extracts some kind of overview of the history of a representative subset, perhaps all, of a series, up to now. However, if the number of time series is large, say hundreds or more, it is not feasible to concatenate them all to form a single input. We reduce the severity of this problem by using a context model that processes just one single series at a time, however within the batch structure, for purposes of efficiency, context series are processed in parallel. At the end of each step through time, we flatten the batch output into one vector, i.e. context vector $\mathbf{r}$. 

Forecasting follows the usual stages of on-the-fly pre-processing (see \cite{Smy21, Smy22}), that includes adaptive normalization and deseasonalization, but then, before reaching the RNN, the input is augmented (concatenated) with the output of the context track, which can be the same for each series in a batch being forecasted. However, if the number of series is relatively small, which means each series can be presented to the forecasting network many, say more than 20, times during training, we can improve the accuracy by adding per-series parameters, i.e. vector $\mathbf{g}$ of the same length as of the context vector $\mathbf{r}$, initialized as ones. Vector $\mathbf{g}$ will modulate the general context information in a way to match particular series needs. This vector is not dynamic, i.e. does not change while stepping through the batch time steps.
If the number of series were larger, which means it would be difficult to fit per-series parameters, we could try to calculate vector $\mathbf{g}$ dynamically, using some kind of attention mechanism, but in our preliminary exploration this resulted in worse forecasting accuracy, and in this work, we do not report further on this possibility.

Both context and forecasting tracks are synchronized in their steps through time. All the parameters, of the context track and main track, and also per-series parameters, if used, are fitted in order to lower the forecasting error. Note that there is no separate loss function applied to the output of the context track.

A batching mechanism allows us to use a relatively large number of series as context series, but of course there are limitations on the size of one batch. One way to reduce the number of context series is to use business knowledge and choose a set representing well all the series (and to make things easier, without missing data). Another way would be to create a factor model, where series are modeled as a linear combination of some limited number of base series, and then use the bases as context series.

\subsection{Model Architecture}

Both tracks, context and main, are based on ES-adRNN, i.e. a hybrid model combining ES with RNN, which is equipped with attentive dilated cells (adRNNcells) \cite{Smy22}. 
The ES-adRNN architecture includes four components: ES component, preprocessing component, RNN component and postprocessing component, which is only used in the main track.

\begin{figure}
\centering
	\includegraphics[width=0.48\textwidth]{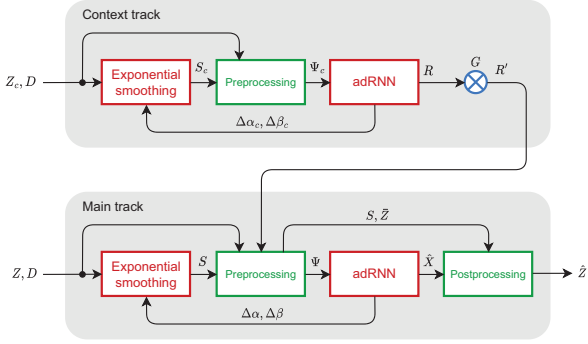}
 \caption{Block diagram of the proposed forecasting model.} 
 \label{figBD}
\end{figure}

The time series denoted in Fig. \ref{figBD} as $Z_c$ (context time series) and $Z$ (forecasted time series) are decomposed individually by ES into seasonal and level components. The seasonal components, $S$, are used by the preprocessing component to deseasonalize time series. Generally, run-time preprocessing prepares training sets for RNN in the context track and main track, $\Psi_c$ and $\Psi$, respectively. The training sets include preprocessed time series and calendar variables $D$. Additionally, training set $\Psi$ includes modulated context vectors $R'$. 

RNN in the main track learns the function mapping input vectors into output patterns which are composed of three parts: (i) vector of encoded point forecasts of hourly loads for the next day, (ii) two vectors encoding predictive intervals (PIs), and (iii) the corrections of ES smoothing parameters, $\Delta\alpha$ and $\Delta\beta$. RNN in the context track, based on input vectors of similar structure to those in the main track except the context part, learns the context vectors $R$. These vectors are modulated by vectors $G$ to adjust them to the individual time series in the main batch.      
A postprocessing component in the main track transforms encoded vectors of point forecasts and PIs, $\hat{X}$, into real values, $\hat{Z}$. To do so, it uses seasonal components $S$ and average values of input sequences $\bar{Z}$ determined by the ES and preprocessing components.

\subsection{Exponential Smoothing Component}

ES component is a simplified Holt-Winters model with multiplicative seasonality. It is composed of two equations expressing a level and weekly seasonality \cite{Smy22}:

\begin{equation}
\begin{aligned}
l_{t,\tau}=\alpha_t \frac{z_\tau}{s_{t,\tau}} + (1-\alpha_t)l_{t,\tau-1} \\
s_{t,\tau+168}=\beta_t \frac{z_\tau}{l_{t,\tau}} + (1-\beta_t)s_{t,\tau}
\label{eqls1}
\end{aligned}
\end{equation} 
where $l_{t,\tau}$ is a level component, $s_{t,\tau}$ is a weekly seasonal component, and $\alpha_t$, $\beta_t \in [0, 1]$ are smoothing coefficients, all for recursive step $t$.

The unique feature of model \eqref{eqls1} is that the smoothing coefficients are not constant as in the standard solution, but they change at each step $t$, adapting dynamically to the current time series properties. The smoothing coefficients are adapted using corrections $\Delta\alpha_t$ and $\Delta\beta_t$ learned by RNN as follows \cite{Smy22}:

\begin{equation}
\begin{aligned}
\alpha_{t+1} = \sigma(I\alpha + \Delta\alpha_t)\\
\beta_{t+1} = \sigma(I\beta + \Delta\beta_t)
\label{eqab}
\end{aligned}
\end{equation} 
where $I\alpha$, $I\beta$ are initial values of the smoothing coefficients, and $\sigma$ is a sigmoid function, which maintains the coefficients within a range from 0 to 1. 

\subsection{Preprocessing and Postprocessing Components}

The goal of preprocessing is twofold. First, it converts the time series to a form that is more convenient for processing by RNN. Second, it creates the input and output patterns and composes them into training sets. 

The primary input for both the context track and main track is the weekly sequence of series preceding directly the forecasted daily period. Let $\Delta^{in}_t$ of size 168 hours be the window covering the $t$-th input weekly sequence, and $\Delta^{out}_t$ of size 24 hours be the window covering the following daily sequence. The input sequence is deseasonalized, normalized and squashed as follows:
\begin{equation}
x_\tau^{in}=\log{\frac{z_\tau}{\bar{z}_t \hat{s}_{t,\tau}} }
\label{eqxt}
\end{equation}where $\tau \in \Delta^{in}_t$, $\bar{z}_t$ is the average value of the input sequence, and $\hat{s}_{t,\tau}$ is the seasonal component predicted by ES for $\tau$ in step $t$.

The preprocessed input sequence is represented by vector $\textbf{x}_t^{in}=[x_\tau^{in}]_{\tau \in \Delta^{in}_t} \in \mathbb{R}^{168}$. The purpose of squashing with a $\log$ function is to prevent outliers from disrupting the learning process. Deseasonalization using $s_{t,\tau}$ removes weekly seasonality while normalization using  $\bar{z}_t$ removes the long-term trend in the input window. Thus all preprocessed series are brought to the same level, which is very important when cross-learning is applied. Note that input patterns $\textbf{x}_t^{in}$ have a dynamic character. They are adapted in each training epoch due to the adaptation of seasonal component $s_{t,\tau}$. This can be seen as the learning of the most useful representation for RNN. 

To enrich the input information, the input patterns for the main track are extended with the level and seasonality of the series, calendar data as well as modulated context vector:
\begin{equation}
\textbf{x}_t^{in'}= [\textbf{x}_t^{in},\, \hat{\textbf{s}}_t,\, \log_{10}(\bar{z}_t),\, \textbf{d}_t^{w},\, \textbf{d}_t^{m},\, \textbf{d}_t^{y}, \textbf{r}'_t] 
\label{eqxp}
\end{equation}where 
$\hat{\textbf{s}}_t \in \mathbb{R}^{24}$ is a vector of 24 seasonal components predicted by ES for $\tau \in \Delta^{out}_t$ reduced by 1, $\textbf{d}_t^{w} \in \{0, 1\}^7, \textbf{d}_t^{m} \in \{0, 1\}^{31}$ and $\textbf{d}_t^{y} \in \{0, 1\}^{52}$ are binary one-hot vectors encoding day of the week, day of the month and week of the year for the forecasted day, respectively, and $\textbf{r}'_t$ is a modulated context vector from the context track (see Section II G).

\begin{figure}
\centering
	\includegraphics[width=0.48\textwidth]{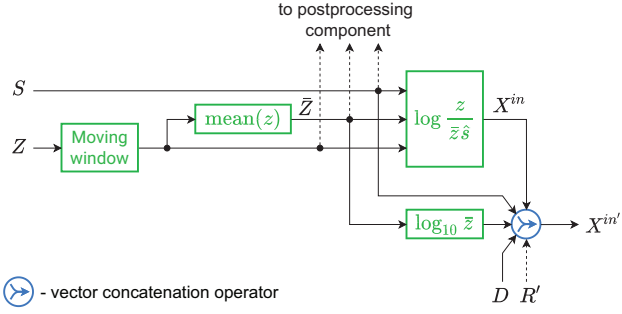}
	\caption{Preprocessing of the input data and generating the input pattern (the links shown with dashed lines do not exist in the preprocessing component for the context track).} 
	\label{figPC}
\end{figure}

An enriched input pattern for the context track has a similar structure to \eqref{eqxp} except for the the modulated context vector.  Calendar variables, $\textbf{d}_t^{w}$, $\textbf{d}_t^{m}$, $\textbf{d}_t^{y}$, carry information about the location of the forecasted daily sequence in the weekly and yearly cycles and help to deal with fixed-date public holidays. Component $\log_{10}(\bar{z}_t)$ informs about the local level of the series, while $\hat{\textbf{s}}_t$ expresses the daily variability of the forecasted sequence.  The modulated context vector, $\textbf{r}'_t$, introduces additional information from the context series for step $t$ adjusted to the forecasted time series. Fig. \ref{figPC} shows the preprocessing of the input data and generating the input pattern. 

The output pattern represents the target daily sequence covered by  $\Delta^{out}_t$. To obtain the output pattern,  $\textbf{x}_t^{out}=[x_\tau^{out}]_{\tau \in \Delta^{out}_t} \in \mathbb{R}^{24}$, the original sequence is normalized as follows:
\begin{equation}
x_\tau^{out}=\frac{z_\tau}{\bar{z}_t }
\label{eqxt2}
\end{equation}where $\tau \in \Delta^{out}_t$.

Note that to define output patterns, we only use normalization so that the errors determined by the loss function for different time series (of different levels) are comparable and have a similar impact on learning. The output patterns are defined only for time series processed in the main track (the context track produces context vectors $\mathbf{r}$, which have no target values). As targets, they are compared with the patterns predicted by the main track and the resulting error is used by the optimization algorithm to adapt all model parameters including both main and context tracks parameters, i.e. ES and RNN parameters as well as modulation vectors $\mathbf{g}$. 

The model is trained on the training patterns produced by shifting by 24 hours the adjacent moving windows, $\Delta^{in}_t$ and $\Delta^{out}_t$. RNN in the main track predicts a vector corresponding to the 24 point forecasts for the next day: ${\hat{\textbf{x}}}_t^{RNN}=[\hat{x}_\tau^{RNN}]_{\tau \in \Delta^{out}_t} \in \mathbb{R}^{24}$. The components of this vector are converted by the postprocessing module into real values using transformed equation \eqref{eqxt}:
\begin{equation}
\hat{z}_\tau=\exp (\hat{x}_\tau^{RNN}) \bar{z}_t\hat{s}_{t,\tau}
\label{eqzt}
\end{equation} where $\tau \in \Delta^{out}_t$.

However, the loss function uses its normalized version, matching \eqref{eqxt2}:
\begin{equation}
\hat{x}^{out}_{\tau}=\frac{\hat{z}_\tau}{\bar{z}_t }
\label{eqxt2.2}
\end{equation}

RNN also predicts two vectors representing bounds of PI: $\hat{\underline{\textbf{x}}}_t^{RNN}$ and  $\hat{\bar{\textbf{x}}}_t^{RNN}$. They are converted into real values in the same way as the point forecasts, using \eqref{eqzt}.

\subsection{RNN Component}
\begin{figure}
	\centering
    \includegraphics[width=0.3\textwidth]{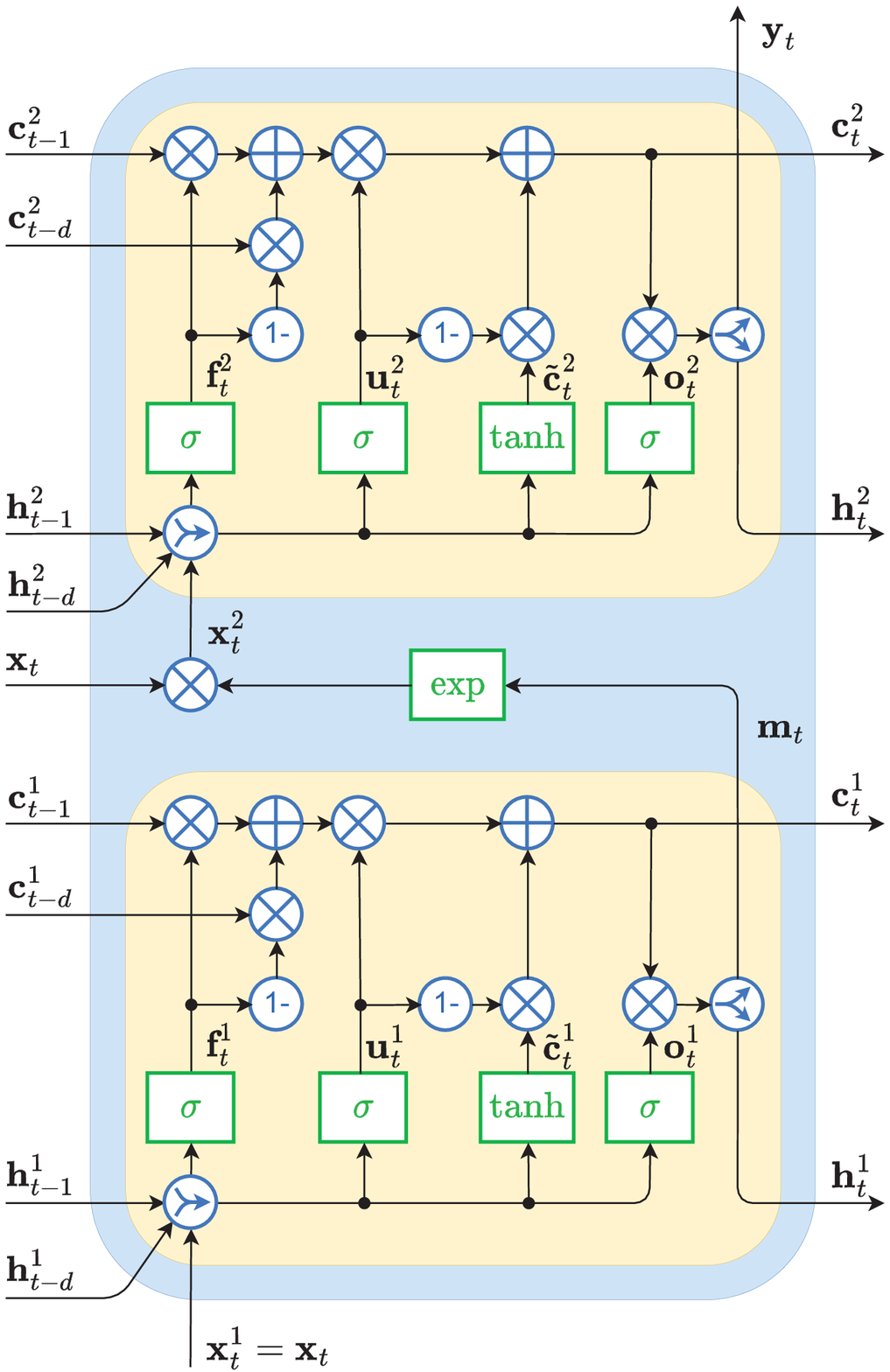}
    \includegraphics[width=0.35\textwidth]{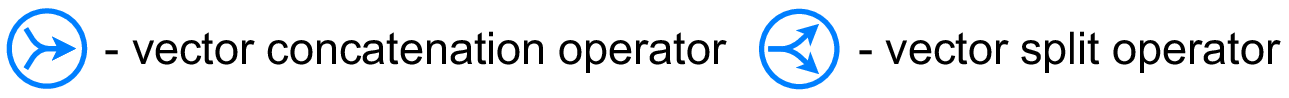}
    \includegraphics[width=0.45\textwidth]{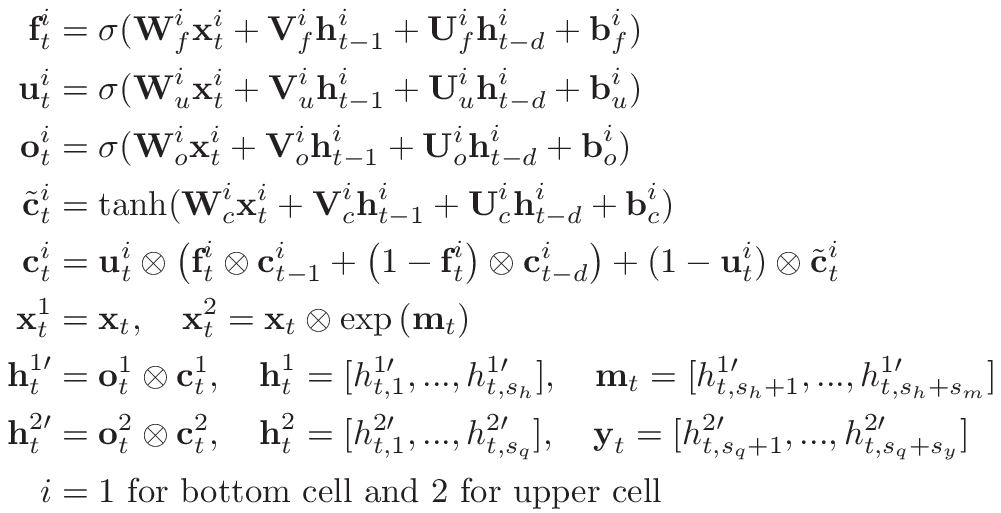}
    \caption{adRNNCell.} 
    \label{figCell3}
\end{figure}

RNN uses our previously proposed LSTM-like cell, adRNNCell, described in \cite{Smy22}. It was designed to deal with multiple seasonality in time series and is equipped with a built-in attention mechanism for weighting the input information. adRNNcell is composed of two identical dRNNCells as shown in Fig. \ref{figCell3}. 

The characteristic features of adRNNCell can be summarized as follows:
\begin{itemize}
    \item Each dRNNCell has two cell states ($c$-states) and two controlling states ($h$-states). The recent states, $\textbf{c}_{t-1}$ and $\textbf{h}_{t-1}$, introduce information to the cell from the recent step $t-1$, as in the standard LSTM solution \cite{Hoch97}, while the dilated states, $\textbf{c}_{t-d}$ and $\textbf{h}_{t-d}$, introduce information from the delayed step $t-d$, $d>1$. These delayed states extend the "receptive field" of the cell which facilitates modeling long-term and seasonal dependencies.
    \item dRNNCell is equipped with two weighting mechanisms, derived from GRU \cite{Cho14}, which manage the $c$-state. The first one, controlled by the $f$-gate, weights recent and delayed $c$-states, $\textbf{c}_{t-1}$ and $\textbf{c}_{t-d}$, while the second one, controlled by the $u$-gate, weights old and new $c$-states, i.e. combination of $\textbf{c}_{t-1}$ with $\textbf{c}_{t-d}$ and $\tilde{\textbf{c}}_t$. Effectively using both recent and delayed states allows the cell to have longer memory, and when used as part of a multi-cell RNN with varying dilations, gives the RNN a multi-resolution characteristics, without the expensive standard attention.
    \item The output of dRNNCell is split into "real output", $\textbf{y}_t$ ($\textbf{m}_t$), which goes to the next layer (cell), and a controlling output, $\textbf{h}_t$, which is an input to the gating mechanism in the following time steps. In LSTM and GRU the functions of both outputs are performed by the $h$-state. In dRNNcell, we split these functions (because they have different tasks to perform) into two different vectors, thus introducing more flexibility. Another idea behind splitting these functions into two outputs is shown in \cite{Ben17}.   
    \item adRNNCell has an internal attention mechanism which performs as follows. The bottom cell produces attention vector $\textbf{m}_t$ (of the same length as input vector $\textbf{x}_t$), whose components, after processing by $\exp$ function, weights the inputs to the upper cell, $\textbf{x}_t$. Based on weighted input vector, $\textbf{x}_t^2$, the upper cell produces vector $\textbf{y}_t$, which feeds the next layer. Note that $\textbf{m}_t$ has a dynamic character. The weights are adjusted by the bottom cell to the current input at step $t$.    
\end{itemize}

The adRNNCell equations at time $t$ are shown in Fig.  \ref{figCell3} where $t$ is a time step;  
$d>1$ is a dilation;
$\textbf{x}_t$ is an input vector; 
$\textbf{h}_{t-1}$ and $\textbf{h}_{t-d}$ are recent and delayed controlling states, respectively;
$\textbf{c}_{t-1}$ and $\textbf{c}_{t-d}$ are recent and delayed cell states, respectively; 
$\textbf{y}_t$ and $\textbf{m}_t$ are outputs of the upper and bottom dRNNCells, respectively;
$\textbf{f}_t$, $\textbf{u}_t$ and $\textbf{o}_t$ are outputs of the fusion, update and output gates, respectively;
$\tilde{\textbf{c}}_t$ is a  candidate state;
$\textbf{W}$, $\textbf{V}$ and $\textbf{U}$ are weight matrices; 
$\textbf{b}$ are bias vectors;
$\sigma$ denotes a logistic sigmoid function;
and $\otimes$ denotes the Hadamard product.

\begin{figure}
	\centering
	\includegraphics[width=0.44\textwidth]{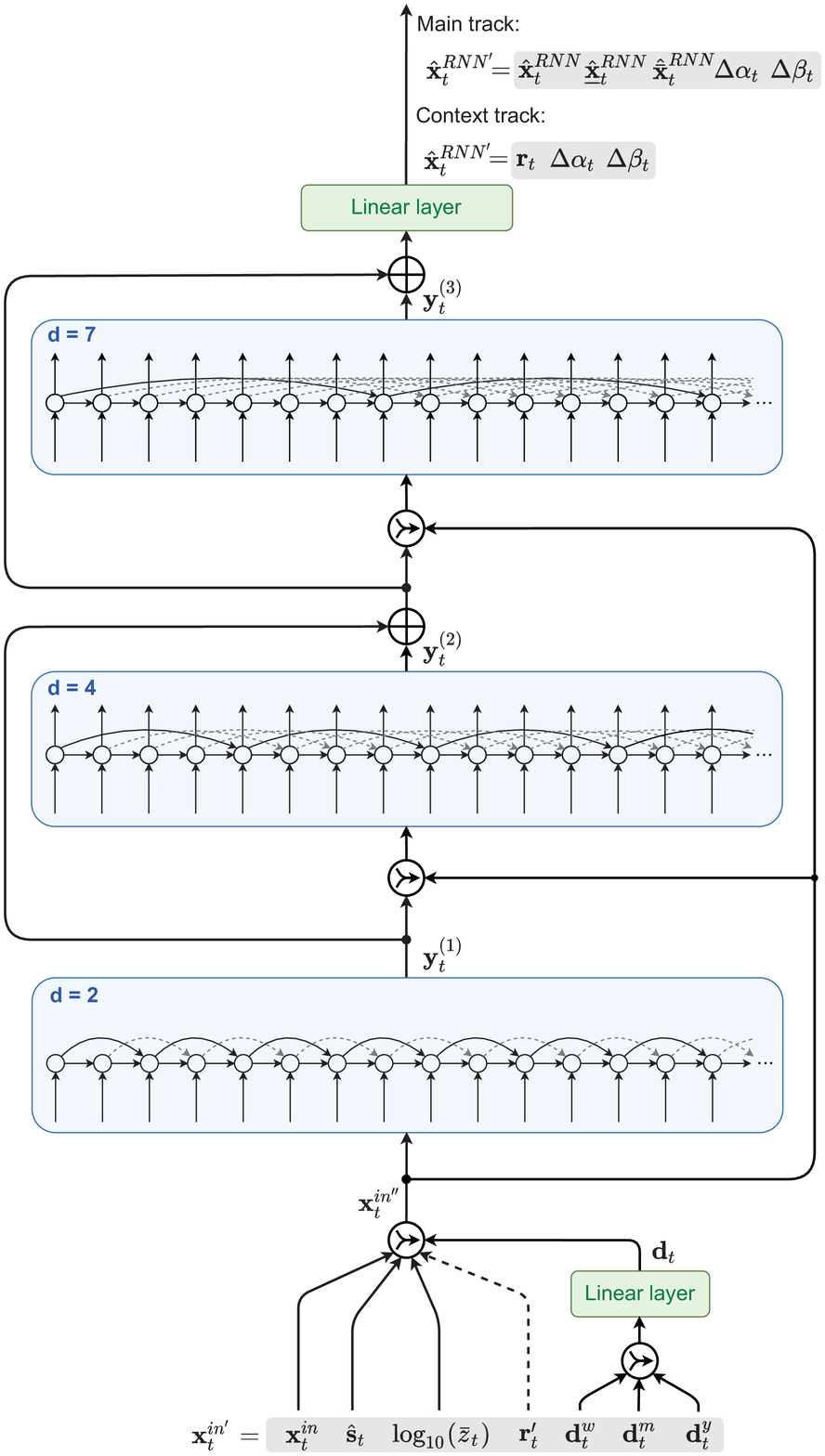}
	\caption{adRNN architecture (the link shown with a dashed line does not exist in the context track).} 
	\label{figRNN}
\end{figure}

Fig. \ref{figRNN} shows  RNN architecture. It is composed of three layers of adRNNCells dilated 2, 4 and 7, respectively. The number of layers as well as the number of dilations were selected by experimentation and may not be optimal for other forecasting problems. 
Stacking several layers with hierarchical dilations helps to extract more abstract features in successive layers and obtain a larger receptive field. This makes it easier to learn the long-term and seasonal temporal dependencies of different scales. However, when the number of stacked cells reaches 3 or more, the gradient may vanish, so we use ResNet-style shortcuts between layers \cite{He16}. Note that input vector $\textbf{x}_t^{in'}$ is introduced not only to the first layer but also to the second and third layers by extending the output vector from the previous layer. 

To reduce the input dimensionality of the calendar variables (90 0-1 slots of $\textbf{d}_t^{w}$, $\textbf{d}_t^{m}$ and $\textbf{d}_t^{y}$), and meaningfully represent sparse binary vectors, a linear layer embeds them  into $d$-dimensional continuous vectors $\textbf{d}_t$. Note that the embedding is learned along with the model itself.

Another linear layer, the output one, produces the adRNN output which for the main track  is composed of five elements: the vector corresponding to the 24 point forecasts,  ${\hat{\textbf{x}}}_t^{RNN}=[\hat{x}_\tau^{RNN}]_{\tau \in \Delta^{out}_t} \in \mathbb{R}^{24}$,  the vector corresponding to the lower bounds of PI, 
 $\hat{\underline{\textbf{x}}}_t^{RNN}=[\hat{\underline{x}}_\tau^{RNN}]_{\tau \in \Delta^{out}_t} \in \mathbb{R}^{24}$, the vector corresponding to the upper bounds of PI, $\hat{\bar{\textbf{x}}}_t^{RNN}=[\hat{\bar{x}}_\tau^{RNN}]_{\tau \in \Delta^{out}_t} \in \mathbb{R}^{24}$, and corrections for smoothing coefficients, $\Delta{\alpha_t}$ and $\Delta{\beta_t}$:

\begin{equation}
\hat{\textbf{x}}_t^{RNN'}= [\hat{\textbf{x}}_t^{RNN},\, 
\hat{\underline{\textbf{x}}}_t^{RNN},\,
\hat{\bar{\textbf{x}}}_t^{RNN},\,
\Delta{\alpha_t},\,
\Delta{\beta_t}]
\label{eqout}
\end{equation} 
For the context track, the output linear layer produces three components: the context vector for the $i$-th context series $\mathbf{r}_{t}^{(i)} \in \mathbb{R}^{u}$, and corrections for smoothing coefficients, $\Delta{\alpha_t}$ and $\Delta{\beta_t}$:

\begin{equation}
\hat{\textbf{x}}_t^{RNN'}= [\textbf{r}_{t}^{(i)}, 
\Delta{\alpha_t},\,
\Delta{\beta_t}]
\label{eqout2}
\end{equation} 
Please note that the $\Delta{\alpha_t}$ and $\Delta{\beta_t}$ in the two equations above, are different, as they are produced by two different RNNs and they are per series.
For $K$ series in the context batch, context RNN produces $K$ vectors $\textbf{r}_{t}^{(i)}$, which are then concatenated, $\textbf{r}_{t}=[\textbf{r}_{t}^{(1)}, ..., \textbf{r}_{t}^{(K)}] \in \mathbb{R}^{uK}$, and modulated by vectors $\textbf{g}_{t}^{(j)}$ (for each series in the main batch, separate $j$-th $g$-vector is adjusted). They extend the input patterns for the main track, see \eqref{eqxp}. We expect that the amount of information extracted from each context series to support the forecasting process in the main track is a monotonic function of the size of the individual context vectors, $u$, although in practice the impact (improvement in accuracy) flattens quickly.

\subsection{Modulation of the Context Vector}

Context vector $\textbf{r}_{t}$ is the same for every series of the batch in the main track. This is not ideal, although of course due to the complexity of the model and its being recurrent, a different response to the same context information for each series will occur. To make this customization stronger, we apply per time series parameters collected in vectors of the same length as the context vector, $\textbf{g}_{t}^{(j)} \in \mathbb{R}^{uK}$, where $j=1, ..., J$, and $J$ is the size of the main batch. Modulation vectors $\textbf{g}_{t}^{(j)}$ initialized as ones are used to element-wise multiply the context, so that each series receives a differently modified context:

\begin{equation}
\textbf{r}_t'^{(j)} = \textbf{r}_t   \otimes  \textbf{g}_t^{(j)}
\label{eqg}
\end{equation}

Modulation vectors $\textbf{g}_{t}^{(j)}$ are 
updated by the same overall optimization procedure
(stochastic gradient descent) as RNN weights, with
the overarching goal of minimizing the loss function.

\subsection{Loss Function}

The model generates point forecasts as well as the lower and upper bounds of PI. To enable the model to optimize these predictions we employ the following loss function \cite{Smy21}:

\begin{equation}
L_\tau =
\rho(x^{out}_\tau, \hat{x}^{out}_{q^*,\tau}) + \gamma(
\rho (x^{out}_\tau, \hat{x}^{out}_{\underline{q},\tau}) + 
\rho (x^{out}_\tau, \hat{x}^{out}_{\overline{q},\tau}))
\label{eqlss}
\end{equation}
where 

\begin{equation}
\rho(x, \hat{x}_q) =
\begin{cases}
(x-\hat{x}_q)q       & \text{if } x \geq \hat{x}_q\\
(x-\hat{x}_q)(q-1)  &\text{if } x < \hat{x}_q 
\end{cases}
\label{eqrho}
\end{equation}
is a pinball loss function defined for $x$ as an actual value, $\hat{x}_q$ as a forecasted value of $q$-th quantile, and $q \in (0, 1)$ as a quantile order; $\tau \in \Delta^{out}_t$; $q^*=0.5$ corresponds to the median; $x^{out}_\tau$ is a normalized actual value \eqref{eqxt2}; $\hat{x}^{out}_{q^*,\tau} = \exp(\hat{x}^{RNN}_\tau)\hat{s}_{t,\tau}$ is a forecasted value of $x^{out}_\tau$; 
$\underline{q}, \overline{q}$ are the quantile orders for the lower and upper bounds of PI, respectively (e.g. $\underline{q}=0.05, \overline{q}=0.95$); 
$\hat{x}^{out}_{\underline{q},\tau} = \exp(\hat{\underline{x}}^{RNN}_\tau)\hat{s}_{t,\tau}$  is a forecasted value of $\underline{q}$-quantile of $x^{out}_\tau$;
$\hat{x}^{out}_{\overline{q},\tau} = \exp(\hat{\bar{x}}^{RNN}_\tau)\hat{s}_{t,\tau}$ is a forecasted value of $\overline{q}$-quantile of $x^{out}_\tau$; and $\gamma \geq 0$ is a parameter controlling the impact of the PI component on the loss function (typically between 0.1 and 0.5). 

Note that in cross-learning, the loss function evaluates the forecasts of multiple time series having different levels and hence different error ranges. In order to unify the impact of these errors on the learning process, we calculate the loss for the normalized values. 

Loss function \eqref{eqlss} has three components. The first one, $\rho(x^{out}_\tau, \hat{x}^{out}_{q^*,\tau})$, which is symmetric for $q^*=0.5$, corresponds to the point forecast. The second and third components, $\rho (x^{out}_\tau, \hat{x}^{out}_{\underline{q},\tau})$ and $\rho (x^{out}_\tau, \hat{x}^{out}_{\overline{q},\tau})$, are related to PI bounds. The pinball loss for them is asymmetric to an extent that depends on the quantile orders. Due to the three-component loss function, both point forecasts and PIs can be simultaneously optimized. Hyperparameter $\gamma$ decides which component is more important in this optimization. For  $\gamma = 1$, all components have the same importance. To increase the importance of the first component, we decrease $\gamma$ towards the limit of 0.

It is worth noting that the pinball loss gives the model the opportunity to reduce the forecast bias by penalizing positive and negative deviations differently. 
When the model tends to bias positively or negatively, we can reduce this bias by entering $ q ^ * $ less than or greater than $ 0.5 $, respectively (see \cite{Smy20, Dud21}). In the same way, we can reduce the bias in PI.

\section{Experimental Study}

In this section we evaluate our proposed model on STLF tasks for 35 European countries. We characterize the data, then we describe the training and optimization procedures, and the baseline models. Next we present the results, the ablation study and finish by discussing the results and model properties. 
\subsection{Data}

The real-world data was collected from ENTSO-E repository (\url{www.entsoe.eu/data/power-stats}). The dataset comprises hourly electricity demand for 35 European countries (we share this data with the model code in our github repository \cite{rep22}). The dataset provides a variety of times series with different properties such as the level and trend, stability in variance over time, intensity and regularity of seasonal fluctuations of different periods (annual, weekly and daily), and intensity of random fluctuations. Diversified time series with different properties allow for a more reliable evaluation of the model.

\begin{figure}
	\centering
	\includegraphics[width=0.45\textwidth]{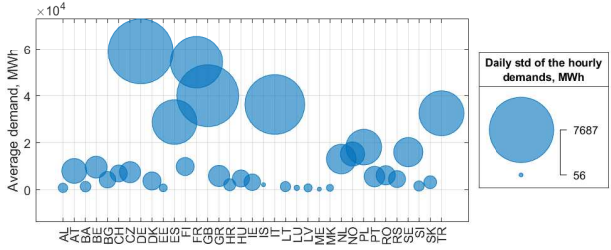}
	\caption{Average demand and daily dispersion of demands for European countries.} 
	\label{figB}
\end{figure}

To show the difference in time series, we include two figures. Fig. \ref{figB} shows the average demand and daily dispersion of the hourly demands for the countries under consideration in the three-year period 2016-18. The average demand varies from 383 MWh for Montenegro (ME) to 58,941 MWh for Germany (DE), while the daily standard deviation of the hourly demands varies from 56 MWh (Iceland, IS) to 7,687 MWh (DE). 

\begin{figure}
	\centering
	\includegraphics[width=0.45\textwidth]{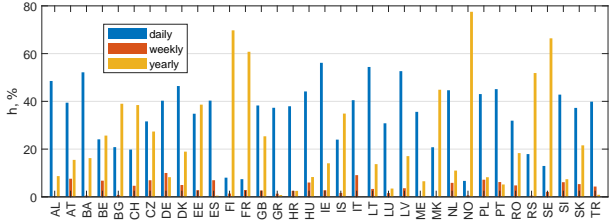}
	\caption{$h$-ratio for daily, weekly and yearly harmonics.} 
	\label{figH}
\end{figure}

Fig. \ref{figH} shows the share of the three most significant seasonalities in the time series. This share is expressed by the ratio $h_i=100 A_i^2/(2Var(z_t))$, where $A_i$ is the $i$-th harmonic amplitude (daily, weekly or yearly) and $Var(z_t)$ is the series variance. As can be seen from Fig. \ref{figH}, for most countries (24) the daily seasonality dominates over others (this is most evident for Bosnia and Herzegovina, Ireland and Lithuania), while for 11 countries the yearly seasonality dominates  (Finland, France, Norway and Sweden are in the lead here). The weekly seasonality is less significant with the $h$-ratio below 10\%. Note that countries similar in terms of average demand and its variance can differ substantially in terms of seasonal components -- compare Great Britain (GB) and Italy (IT). In the case of the latter, Fourier analysis does not reveal the yearly seasonality. 

\subsection{Training, Optimization and Evaluation Setup}

The data time span is from 2006 to 2018 but a lot of data is missing in this period. Complete time series, without missing values, were selected to $Z_c$ as context series. They include series for 18 countries: AT, BA, BE, BG, CZ, DE, ES, FR, HR, HU, LU, MK, NL, PL, PT, RO, SI, and SK.

The model development and hyperparameter search was executed by splitting the data into training (years 2006-2015) and validation (2016-2017) sets. Generally, the hyperparameters were chosen so as to minimize the forecasting validation error, however, only after achieving close to zero bias of the forecast, by manipulating $ q ^ * $ -- see discussion of the loss function above.
Then, using the best hyperparameters, a final training was done on data from 2006 to 2017, and the test was performed on 2018 data.
We predict the daily load profiles for each day of the test period, for each of the 35 countries. 
Due to missing values for Estonia and Italy (last month of 2018 is missing) and Latvia (last two months of 2018 are missing), the final errors do not cover these missing periods. 

Each training epoch is composed of $n_o$ of "sub-epochs", defined in the traditional fashion as one scan of all available data. The number of sub-epochs was determined from \cite{Smy21}:
\begin{equation}
n_o=\min{\left(1,\left(\frac{Nb}{L}\right)^p\right)}
\end{equation}where $N$ is the maximum number of updates per epoch, $b$ is the current batch size, $L$ is the number of time series in the data set, and $p \in [0,1]$ is a hyperparameter, which by experimentation is set to 0.8.

We use the same schedule of increasing batch sizes and decreasing learning rates proposed in \cite{Smi18}. We start with a small batch size of 2, and increase it, although only once, due to the small number of series, to 5 at epoch 4. 
We use another schedule of decreasing learning rates, which has a similar, but not exactly  the same effect as increasing the batch size, this allows the validation error to be  further reduced. We use the following schedule: $3\cdot10^{-3}$ (epochs 1-5), $10^{-3}$ (epoch 6), $3\cdot10^{-4}$ (epoch 7), and $10^{-4}$ (epochs 8-9).

The sizes of $c$-state and $h$-state were  $s_c=150$, $s_h=70$. These values were obtained by experimentation starting with $s_c=50$, $s_h=20$ and increasing it 3 times, by around 50 and 25, respectively).

During each epoch a number of updates is executed, each guided by average error accumulated by executing up to $l_o$ (e.g. 50) forward steps, moving by one day, on a batch. The starting point is chosen randomly and this may cause the number of batch steps be smaller than 50. The batches include random $b$ series. The model is trained using an Adam optimizer.

In \cite{Smy22} we discussed the architecture-related hyper parameters, please refer there for more details. As stated there, using experimentation and some rules of thumb (e.g. one of the dilations should be equal seasonality length) we arrive at three blocks, each composed of one cell, with the following dilations: 2, 4 and 7, see  Fig. \ref{figRNN}.

As described in Section II C, the pinball loss function was utilized, with three different quantile values $q$,  to achieve quantile regression for 0.5, 0.05, and 0.95. The actual values for $q^*$, $\underline{q}$, and $\overline{q}$ were slightly different: 0.525, 0.045, 0.975. These values were arrived at by fine tuning during validation, reducing the bias of the center value, and trying to match the expected percentage of exceedance for 0.05 and 0.95 PIs (5\%).

Size of the context batch $K$ was 18. It included all series without any missing values. This condition simplified the code, and as 18 is a relatively small number, all these series could be processed in one batch. 

The size of the individual context (size of the output of the context RNN) $u$ was 3 and was chosen by experimentation starting from 1. 
Generally it should not be too large, so the $u \cdot K$ should not be much larger than the rest of the combined input into the forecasting RNN. The context is, after all, secondary information.

Finally, we used the ensemble size of $E=100$, although a size as small as 5 is often sufficient. Simple mean aggregation was applied to the forecasts.
For other details see \cite{Smy21}.

\subsection{Baseline Models}

The performance of the proposed model is compared with the baseline models. They include statistical models, classical ML models, as well as recurrent, deep and hybrid NN architectures: 

\begin{itemize}
\item Naive -- naive model in the form: the forecasted demand profile for day $i$ is the same as the profile for day $i-7$
\item ARIMA -- autoregressive integrated moving average model \cite{Dud15},
\item ES -- exponential smoothing model \cite{Dud15},
\item Prophet -- modular additive regression model with nonlinear trend and seasonal components \cite{Tay18},
\item N-WE -- Nadaraya–Watson estimator \cite{Dud15}
\item GRNN -- general regression NN \cite{Dud16a},
\item MLP -- perceptron with a single hidden layer and sigmoid nonlinearities \cite{Dud16a},
\item SVM -- linear epsilon insensitive support vector machine ($\epsilon$-SVM) \cite{Pel21},
\item LSTM -- long short-term memory \cite{Pel20},
\item ANFIS -- adaptive neuro-fuzzy inference system \cite{Pel18},
\item MTGNN -- graph NN for multivariate TS forecasting \cite{Wu20}.
\item DeepAR -- autoregressive RNN model for probabilistic forecasting \cite{Sal20},
\item WaveNet -- autoregressive deep NN model combining causal filters with dilated convolutions \cite{Oor16},
\item N-BEATS -- deep NN with hierarchical doubly residual topology \cite{Ore20},
\item LGBM -- Light Gradient-Boosting Machine \cite{Ke17},
\item XGB -- eXtreme Gradient Boosting algorithm \cite{Che16},
\item ES-adRNNe100 -- hybrid model combining ES and dilated RNN with attention mechanism \cite{Smy22} (predecessor of the proposed model). The result are presented for ensemble of 100 ES-adRNN base models.

\end{itemize}
In our recent works \cite{Smy22} and \cite{Smy21}, the models were trained on the data from the period 2016-18. In this work we extend the training period to 2006-18. But some models achieved better results on the shorter period 2016-18, and some of them were not able to learn on data with missing values (about 40\% of the countries have incomplete data in the period 2006-2015). For these models, which are marked with an asterisk in Table \ref{tabEr}, we provide results for the shorter training period.

\subsection{Results}

\begin{figure}
\centering
	\includegraphics[width=0.28\textwidth]{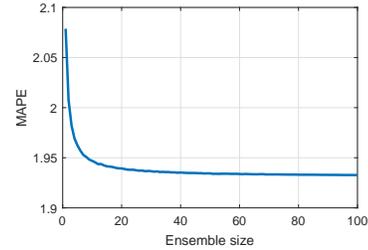}
 \caption{MAPE depending on the ensemble size.} 
 \label{figES}
\end{figure}

\begin{figure}
\centering
	\includegraphics[width=0.46\textwidth]{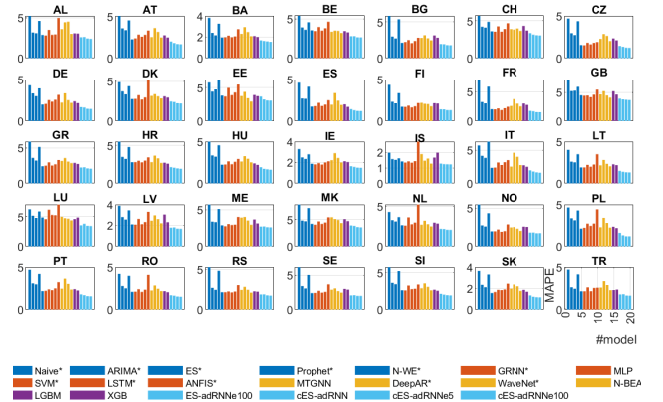}
 \caption{MAPE for individual countries.} 
 \label{figP}
\end{figure}

The forecasting quality metrics are shown in Table \ref{tabEr}. They include: mean absolute percentage error (MAPE), median of APE (MdAPE), interquartile range of APE (IqrAPE), root mean square error (RMSE), mean PE (MPE), and standard deviation of PE (StdPE). We provides the results of our proposed model in three variants: individual model, cES-adRNN, ensamble of five individual models, cES-adRNNe5, and ensemble of 100 individual models, cES-adRNNe100. As can be seen from Table \ref{tabEr}, the hybrid models combining ES and adRNN outperform all other models in terms of MAPE, MdAPE and RMSE. Even in its individual version, cES-adRNN, it produces more accurate forecasts than ensembled models such as N-BEATS and boosted-tree models. How ensembling improves results of cES-adRNN is shown in Fig. \ref{figES}. For over 20 ensemble members, the MAPE stabilizes, reaching 1.93 for 100 members. 
The proposed model with a context track is substantially better in terms of accuracy than its predecessor: compare cES-adRNNe100 and ES-adRNNe100 in Table \ref{tabEr}. MAPE and RMSE were reduced by almost 10\%, while MdAPE was reduced by more than 11\%. Fig. \ref{figP} shows MAPE for individual countries. For all countries cES-adRNNe100 yielded the lowest errors except Estonia (EE), where it was beaten by N-BEATS.

Note the lowest values of IqrAPE and StdPE in Table \ref{tabEr} for the proposed model, which indicates the least dispersed predictions compared to the baseline models. The only category in which our model failed to beat
other models is MPE. MPE reflects a forecast bias. Note that our model can reduce the bias by selecting appropriate quantile order for the loss function. 
But it must be remembered that bias reduction may have a negative impact on the forecast error which is the main quality measure for us. 
This was the reason we did not reduce the bias further, leaving it at $-0.25$.
  
To confirm the best accuracy of the proposed model, a pairwise one-sided Giacomini-White test for conditional predictive ability was performed individually for each country \cite{Gia06}. The results are summarized in Table \ref{tabEr}, where GWtest is a percentage of cases for which a model is significantly better in terms of MAPE than other models at $\alpha=0.05$. GWtest = 95.49 for cES-adRNNe100 means that this model in 95.49\% pairwise comparisons with other models had significantly lower MAPE. 


\begin{table}
	\setlength{\tabcolsep}{2.45pt}
	\caption{Quality metrics.}
	\begin{tabular}{lcccrrcr}
		\toprule
		& MAPE  & MdAPE & IqrAPE & RMSE  & MPE   & StdPE &GWtest\\
		\midrule    
  
    Naive* & 5.08  & 4.84  & 3.32  & 704.34 & -0.26 & 7.91& 1.05\\
    ARIMA* & 3.30  & 3.01  & 3.00  & 475.09 & -0.01 & 5.31& 18.35\\
    ES*   & 3.11  & 2.88  & 2.73  & 439.26 & 0.01  & 5.13& 27.37\\
    Prophet* & 4.53  & 4.32  & 3.03  & 619.39 & -0.13 & 6.82& 6.77\\
    N-WE* & 2.49  & 2.28  & 2.30  & 332.49 & -0.13 & 4.26& 59.25\\
    GRNN* & 2.48  & 2.28  & 2.27  & 332.91 & -0.11 & 4.25& 61.80\\
    MLP   & 2.92  & 2.68  & 2.78  & 395.72 & 0.09  & 5.04& 33.68\\
    SVM*  & 2.55  & 2.29  & 2.52  & 357.24 & -0.13 & 4.37& 56.69\\
    LSTM* & 2.76  & 2.57  & 2.52  & 381.76 & 0.02  & 4.47& 39.70\\
    ANFIS* & 3.65  & 3.17  & 3.66  & 507.08 & -0.10 & 6.43& 13.53\\
    MTGNN & 2.87  & 2.62  & 2.59  & 372.65 & -0.02 & 4.64& 33.68\\
    DeepAR* & 3.42  & 3.25  & 2.95  & 487.14 & -0.51 & 5.16& 17.59\\
    WaveNet* & 3.03  & 2.84  & 2.69  & 417.49 & -0.83 & 4.68& 28.42\\
    N-BEATS* & 2.56  & 2.36  & 2.39  & 356.83 & -0.04 & 4.29& 51.58\\
    LGBM  & 2.87  & 2.60  & 2.52  & 391.16 & -0.05 & 4.64& 36.99\\
    XGB   & 2.69  & 2.43  & 2.42  & 366.97 & \textbf{0.00} & 4.20& 48.57\\
    ES-adRNNe100 & 2.14  & 1.93  & 2.09  & 290.89 & -0.11 & 3.60& 81.95\\
    cES-adRNN & 2.08  & 1.86  & 2.04  & 281.15 & -0.25 & 3.39& 84.66\\
    cES-adRNNe5 & 1.96  & 1.74  & 1.93  & 265.37 & -0.25 & 3.22& 93.23\\
    cES-adRNNe100 & \textbf{1.93} & \textbf{1.71} & \textbf{1.91} & \textbf{262.65} & -0.25 & \textbf{3.19} & \textbf{95.49}\\
		\bottomrule
	\end{tabular}
	\label{tabEr}
\end{table}

\begin{figure}
\centering
	\includegraphics[width=0.43\textwidth]{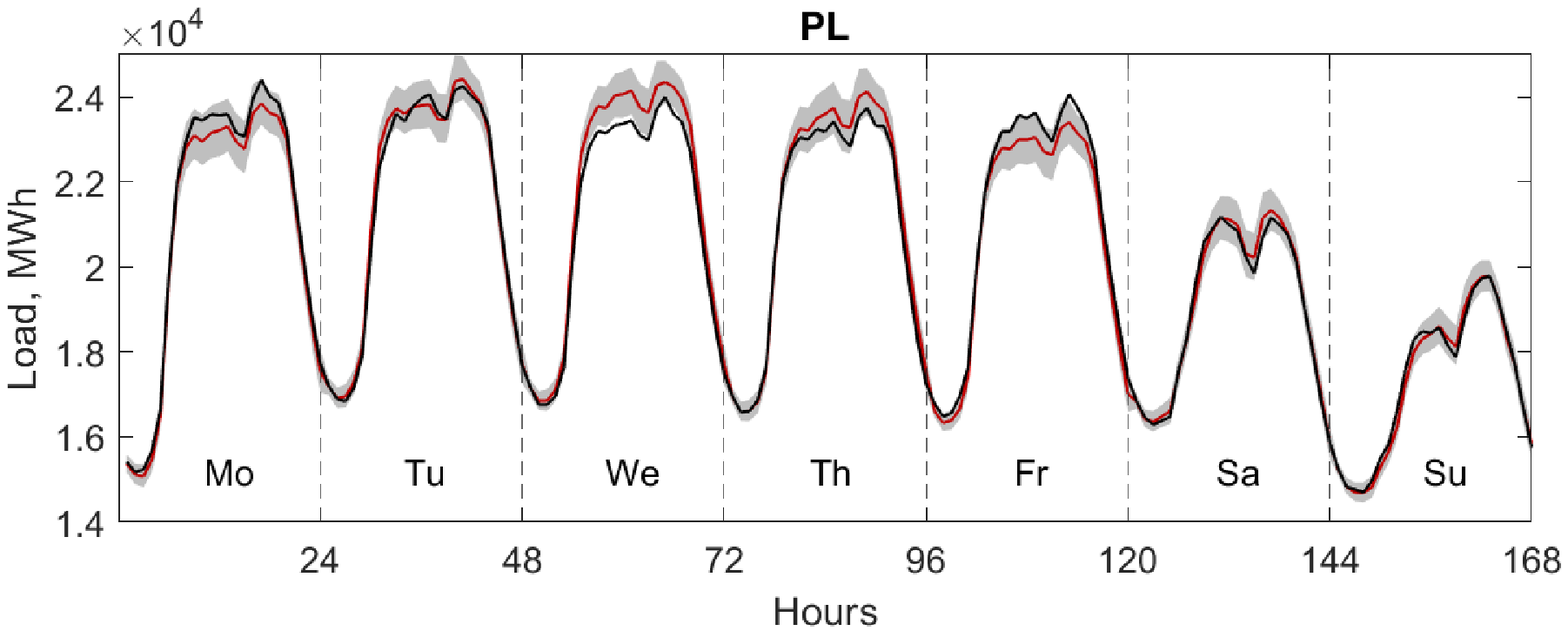}
 \includegraphics[width=0.43\textwidth]{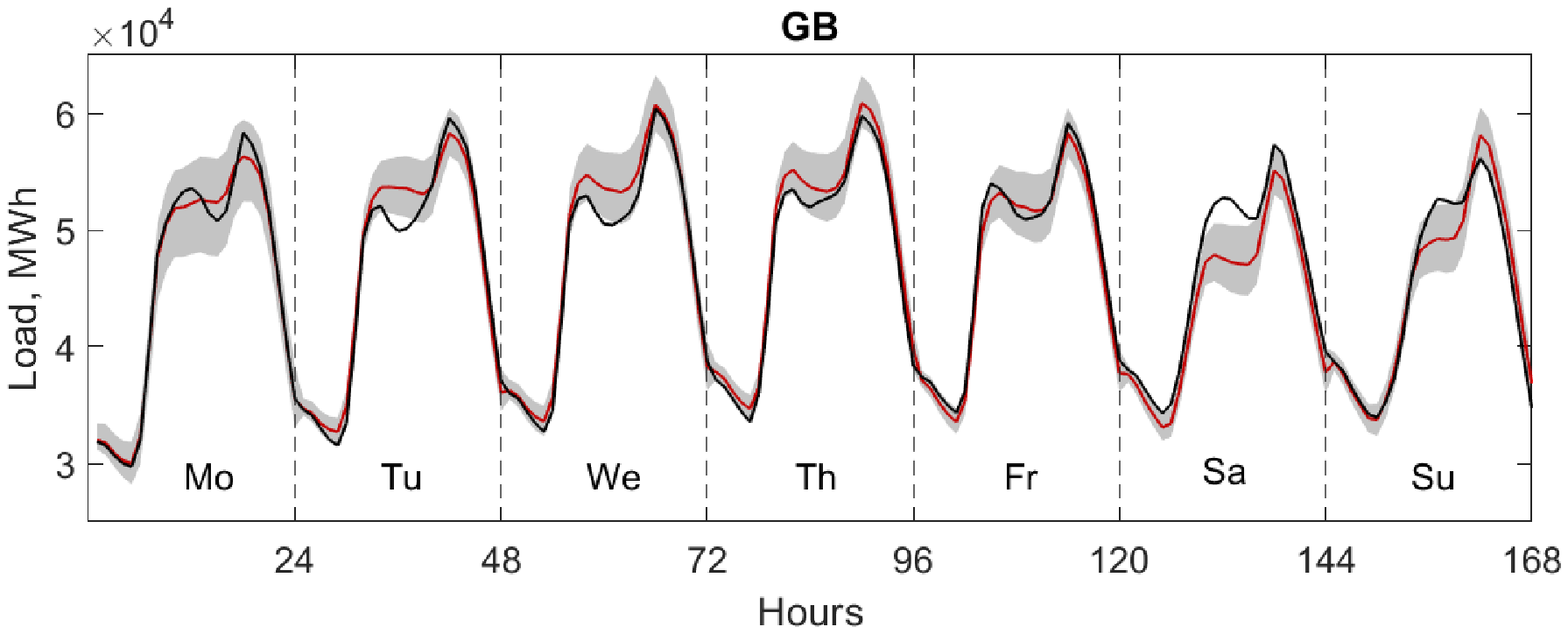}
 \includegraphics[width=0.43\textwidth]{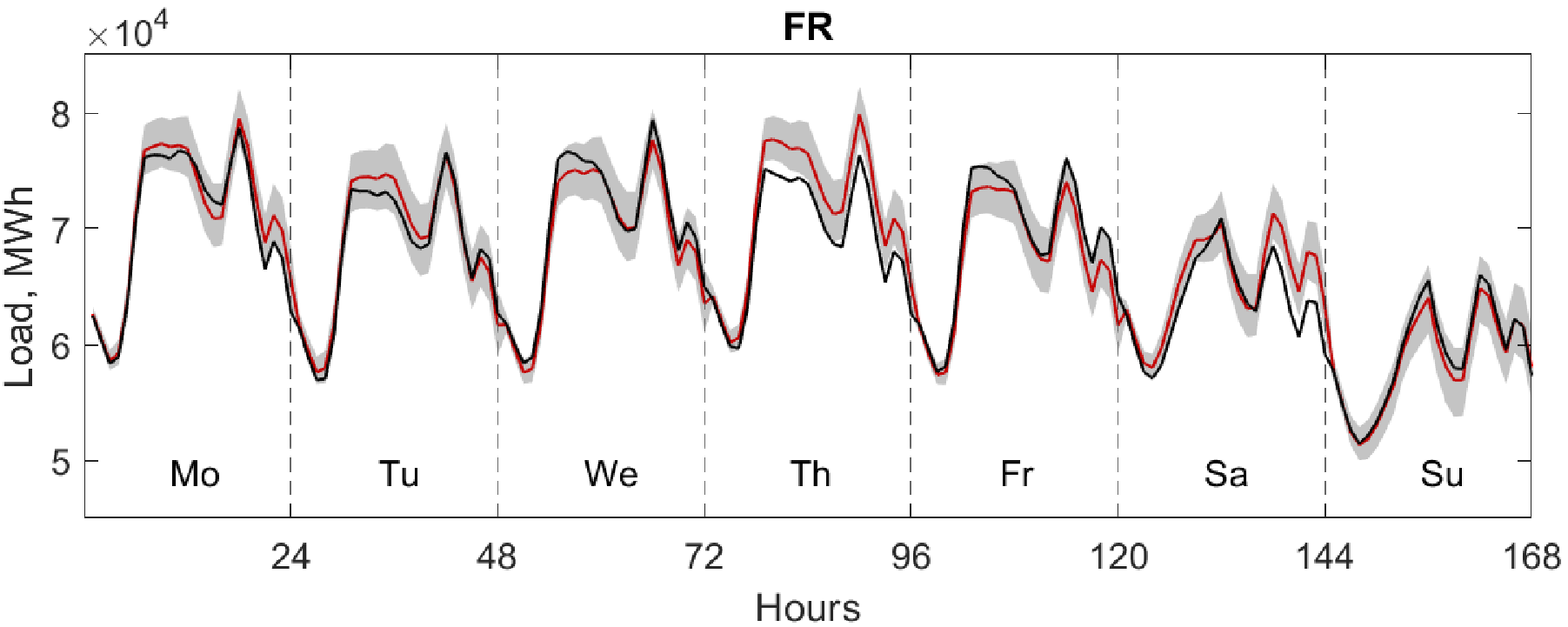}
 \includegraphics[width=0.43\textwidth]{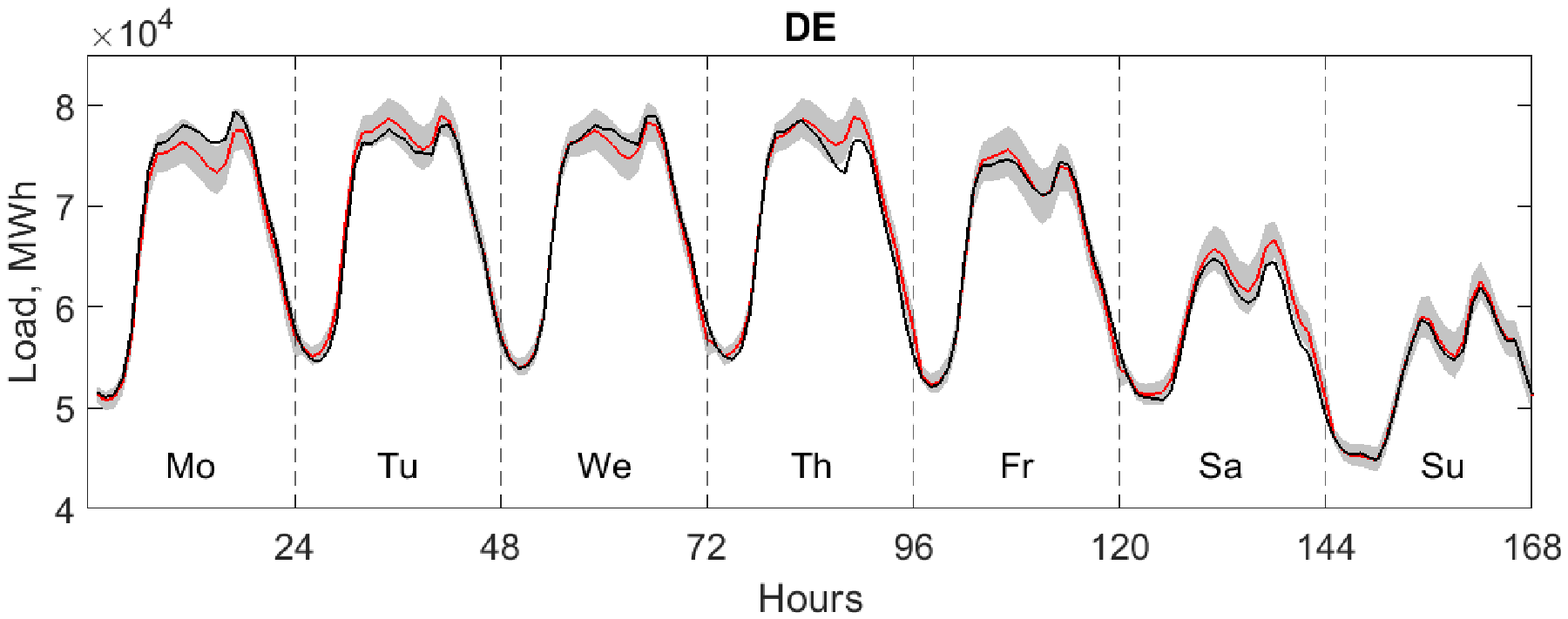}
 \caption{Examples of forecasts produced by cES-adRNNe100 (real in black, forecasts in red, 90\% PIs in gray).} 
 \label{fig8}
\end{figure}

Fig. \ref{fig8} shows examples of forecasts produced by cES-adRNNe100 (weekly period from January 2018). The PIs are also shown in this figure. To evaluate PIs, we calculate for each country the number of observed values in PIs, below PIs and above PIs. The target values for the assumed quantiles are: 90\%, 5\% and 5\%, respectively. We achieved: $90.51\% \pm 3.21\%$, $5.86\% \pm 1.85\%$ and $3.63\% \pm 1.47\%$, respectively. In addition, we calculate the Winkler score which is defined as the length of the PI plus a penalty if the observation is outside the interval. To make Winkler scores for different countries comparable, we divide each score by the average series value for the country in the test period. The mean normalized Winkler score is $0.1229\% \pm 0.2374\%$.

\subsection{Ablation Study}

In this section, we investigate new elements of cES-adRNNe and estimate their influence on the results. 
We compare the results of the full model with results of the reduced model:
\begin{description}
\item[\textbf{Ab1}] ES-adRNNe, without the context track. This is a predecessor of cES-adRNNe described in \cite{Smy22}. Only the main track is used, so no additional inputs from the context track are introduced.  

\item[\textbf{Ab2}] cES-adRNNe without connections introducing input vector $\textbf{x}_t^{in'}$ directly to the second and third layers by extending the output vector from the previous layer (see Fig. \ref{figRNN}). 

\item[\textbf{Ab3}] cES-adRNNe without adaptive modulation vector $\mathbf{g}$. Thus the context vector is the same for all 
series in the batch of the main track.

\item[\textbf{Ab4}] Learning on a shorter period 2016-17. In our previous works on ES-RNN for STLF we used such a period because it does not contain missing data. In this work we extend this period to 2006-17. 
\end{description}

Other mechanisms and components of cES-adRNNe, which influence its performance were investigated in \cite{Smy21} (see Section IVD Ablation Study). They include: ES component, ResNet-style shortcut connections, fusion gate, recent and dilated states, selection of the input data, and embedding of the calendar data. It was shown that all of them are beneficial to the performance of the ES-RNN model for STLF.

Table \ref{tabAS} shows MAPE and RMSE for the full model and its reduced variants (results for ensemble of 5 base learners are shown). As can be seen from this table, the most beneficial action to improve the accuracy of the model is to train it on longer time series, from 2006 to 2017 (Ab4). In this case MAPE was reduced from 2.25 to 1.96. The second most beneficial operation was introducing the context track, which supports learning of the main track (Ab1). This reduced MAPE by 0.18 percentage point. Introducing an adaptive modulation vector adjusted to each series (Ab3) reduced MAPE by 0.07 percentage point. The smallest reduction in error was observed after introducing direct links between the input layer and subsequent recurrent layers (Ab2). 

\begin{table}[htbp]
  \centering
  \caption{Results of ablation study}
    \begin{tabular}{lrrrrrr}
    \toprule
          & \multicolumn{1}{l}{Full} & \multicolumn{1}{l}{Ab1} & \multicolumn{1}{l}{Ab2} & \multicolumn{1}{l}{Ab3} & \multicolumn{1}{l}{Ab4}  \\
    \midrule
    MAPE  & 1.96 & 2.14 & 1.98 & 2.03 & 2.25  \\
    RMSE  & 265 & 292 & 269 & 276 & 307    \\
    \bottomrule
    \end{tabular}%
  \label{tabAS}%
\end{table}%

\subsection{Discussion}

Context information adjusted by a modulation vector to individual time series improves the forecasting accuracy, as shown by the ablation study (Ab1, Ab3). 
Even old information, from a few years ago, was important for forecasting new data (Ab4), although it seems intuitive that it should not have such a significant impact on the current dynamics of the series. By extending the training period to 2006-18, the model was able to reduce MAPE and RMSE by more than 12\%.

The proposed model is trained globally across all time series (cross-learning) to exploit the information dispersed across many series \cite{Jan20}. This is a type of multi-task learning \cite{Car97} based on latent correlation across multiple series \cite{Smy20}. It improves generalization by preventing overfitting to individual series and, in addition, it significantly speeds up training. Although the model is trained globally, this does not mean that it is not sensitive to the individual properties of the time series. 
They are expressed in per-series ES parameters and modulation vectors. The characteristic features of each individual series are extracted dynamically by ES and used for learning an appropriate representation for RNN. 

RNN, due to its multiple dilated stacked architecture based on a new type of cells, adRNNCells, is able to model complex time series with multiple seasonality. The delayed connections of adRNNCells extend their receptive fields and enable them to model not only short-term but also long-term and seasonal relationships. This is done in a hierarchical manner: subsequent layers expand the receptive field more and more. The attention mechanism built into adRNNCells selects the most relevant input information for improving the forecast accuracy. This mechanism is implemented by a recurrent cell (bottom dRNNCell) to make it dynamic. 
In each recurrent cycle, a new attention vector is produced based on the current input vector and historical information accumulated in the cell states.

A hybrid architecture enables the model to be optimised and trained as a whole. All the components of the context track and the main track, i.e. ES components, adRNNs, and modulation vector production are optimized simultaneously by the same optimization algorithm. This entails learning of the series representations in both tracks, learning context vector by the context adRNN, learning a way to produce modulation vector, learning of the smoothing coefficients and initial ES parameters, and learning point forecasts and predictive intervals by the main adRNN in the same time. 

The model produces both point forecasts and predictive intervals for them. This is very useful in practical applications because it gives additional information on how reliable the forecast is. To enable the model to predict point forecasts and predictive intervals we introduced the combined loss function based on pinball loss. Using it, the forecast bias can be also controlled.

To prevent overfitting, the model uses ensembling (in the current version we use the simplest variant of ensembling: averaging forecasts generated in the $E$ independent runs of the model with random initialization, but other more sophisticated ensembling methods are possible, see \cite{Smy20, Dud21}). 
Also, at the time of batch creation, the starting point of training is sampled, and taking into account that the number of steps is limited, each time a series is a member of a batch, a different piece of the series is used. This is an additional mechanism that delays overtraining.

The ResNet-style shortcuts make the system additive: second and third layers add their adjustments to the output of the first layer. Extending input to these two higher layers with $\textbf{x}_t^{in'}$ improves the forecasting performance. Somehow, these additions are more accurate when the higher layers "see" the original input.

\section{Conclusion}

In this study, we propose a novel contextually enhanced hybrid model for STLF combining RNN with attentive dilated cells and exponential smoothing. The context track extracts information from the representative time series and  introduces it as an additional input to the main track, which produces point forecasts and predictive intervals. The experimental study showed that the context track improves significantly forecasting accuracy. The proposed model outperformed in terms of accuracy all baseline models including statistical models, state-of-the-art machine learning models, and its predecessor, which does not have a context track. 

The proposed model is equipped with many mechanisms and procedures designed to increase forecasting efficiency:

\begin{itemize}
\item hybrid architecture – ES for on-the-fly preprocessing (learning of time series representation) and RNN for modeling temporal dependencies in the series,
\item a context RNN that provides additional global information to the main RNN,
\item new recurrent cells with dilation and attention mechanisms, which help in modeling long-term and seasonal dependencies as well as selecting input information,
\item dynamic ES model – RNN corrects ES parameters in each recurrent cycle, 
\item cross-learning, which enables the model to capture shared features of individual series and prevents overfitting,
\item quantile loss function which enables the model to produce both point forecasts and predictive intervals, and also to reduce the forecast bias,
\item ensembling and mechanisms to delay overfitting.
\end{itemize}


%


\ifCLASSOPTIONcaptionsoff
  \newpage
\fi

\end{document}